\documentclass[runningheads]{llncs}

\usepackage[T1]{fontenc}
\usepackage[utf8]{inputenc}
\usepackage{amsmath,amssymb}
\usepackage{booktabs}
\usepackage{multirow}
\usepackage{array}
\usepackage{graphicx}
\usepackage{xcolor}
\usepackage{tikz}
\usepackage{subcaption}
\usepackage{tabularx}
\usepackage{booktabs}
\usepackage{comment}
\usepackage[numbers,sort&compress]{natbib}

\newcolumntype{Y}{>{\raggedright\arraybackslash}X}
\usetikzlibrary{trees,positioning,fit,backgrounds,shapes.geometric,arrows.meta}
\usepackage{pgfplots}
\pgfplotsset{compat=1.18}
\usepackage[hidelinks]{hyperref}

\usepackage[export]{adjustbox}

\newcommand{\rev}[1]{\textcolor{black}{#1}}  

\pgfdeclarelayer{edgelayer}\pgfdeclarelayer{nodelayer}
\pgfsetlayers{background,edgelayer,nodelayer,main}

\definecolor{cInput}{RGB}{30,41,59}
\definecolor{cNeural}{RGB}{217,119,6}
\definecolor{cLogical}{RGB}{37,99,235}
\definecolor{cGating}{RGB}{22,163,74}

\newcommand{\inputcolour}[1]{\textcolor{cInput}{#1}}
\newcommand{\neuralcolour}[1]{\textcolor{cNeural}{#1}}
\newcommand{\logicalcolour}[1]{\textcolor{cLogical}{#1}}
\newcommand{\gatingcolour}[1]{\textcolor{cGating}{#1}}
\newcommand{\neural}{\mathcal{N}}
\newcommand{\logical}{\mathcal{L}}
\newcommand{\gating}{G}
\newcommand{\probneural}{P_{\neural}}
\newcommand{\problogical}{P_{\logical}}
\newcommand{\concordia}{C}

\tikzset{
  none/.style        ={inner sep=0pt, minimum size=0pt},
  black/.style       ={draw=black, thick},
  smallgreen/.style  ={circle, draw=cGating, fill=cGating!12,
                       inner sep=1pt, minimum size=4.5mm},
  orangesolid/.style ={draw=cNeural,  thick},
  bluesolid/.style   ={draw=cLogical, thick},
  greensolid/.style  ={draw=cGating,  thick},
  orangedashed/.style={draw=cNeural,  thick, dashed},
  orangedirected/.style={draw=cNeural,  thick, -{Latex[length=2mm]}},
  bluedirected/.style  ={draw=cLogical, thick, -{Latex[length=2mm]}},
  greendirected/.style ={draw=cGating,  thick, -{Latex[length=2mm]}},
}

\definecolor{cBoolean}{HTML}{2C3E50}
\definecolor{cProb}{HTML}{2980B9}
\definecolor{cDiff}{HTML}{27AE60}
\definecolor{cFuzzy}{HTML}{E67E22}
\definecolor{cEmbed}{HTML}{8E44AD}
\definecolor{cBox}{HTML}{ECF0F1}

\newcommand{\system}[1]{\textsf{#1}}
\newcommand{\cmark}{\ensuremath{\checkmark}}
\newcommand{\bmark}{\ensuremath{(\checkmark)}}

\begin{document}

\title{Rule-Based Languages for Neurosymbolic AI}
\titlerunning{Rule-Based Languages for Neurosymbolic AI}

\author{Stefania Dumbrava\inst{1}\orcidID{0000-0002-6664-0620} \and\\
Efthymia Tsamoura\inst{2}\orcidID{0009-0008-8302-1902}}
\authorrunning{S. Dumbrava and E. Tsamoura}

\institute{ENSIIE, SAMOVAR, Télécom SudParis, Inria, IRIF, France\\
\email{stefania.dumbrava@ensiie.fr} \and
Huawei Labs, United Kingdom\\
\email{efthymia.tsamoura@huawei.com}}

\maketitle

\begin{abstract}
Logic programming is increasingly used as the symbolic component of neurosymbolic AI systems. We survey the main rule-based languages in this setting, namely Datalog, answer set, and probabilistic logic programs, along four axes: semantics, expressiveness, neural integration, and evaluation mechanism. We analyse over 50 recent systems and applications, comparing formalism usage across four research areas: databases and programming languages, machine learning, vision, and robotics. We provide a decision matrix mapping application scenarios to required features and close by outlining open problems.
\end{abstract}

\section{Introduction}
Recent progress in deep learning and large language models (LLMs) has driven what
Kautz~\cite{kautz2022} calls the ``third AI summer''. LLMs, however,
remain logically inconsistent, rely on shortcut features rather than robust generalisation~\cite{yuan2024shortcut}, and offer no formal guarantees. Neurosymbolic
(NeSy) AI addresses these weaknesses by coupling neural learning with
symbolic reasoning~\cite{marra2024survey,desmet2025defining}. For the
symbolic component, logic programming is a well-suited
candidate. Beyond recursion and relational structure, it brings an
evaluation theory with least-fixpoint semantics whose Datalog core has
polynomial-time data
complexity~\cite{DBLP:books/aw/AbiteboulHV95,DBLP:journals/tkde/CeriGT89,DBLP:conf/rweb/Krotzsch25}.

We focus on systems that reason over an explicit rule layer, namely
Datalog and its extensions, answer set programs (ASP), Prolog-style
(probabilistic) logic programs, and the inductive logic-programming
(ILP) methods that learn such rules. Purely sub-symbolic models,
including graph neural networks, are outside the scope of this work. Our survey is
structured around two aspects of the rule layer. We identify the
language features a task requires, such as recursion, negation,
aggregation, probabilistic semantics, or soft unification, together
with the evaluation mechanism that makes their combination tractable.

State-of-the-art systems extend the base language along orthogonal
directions. Provenance semirings parameterise evaluation, enabling
probabilistic and differentiable
semantics~\cite{DBLP:conf/pods/GreenKT07,DBLP:conf/nips/HuangLCSNSS21,DBLP:journals/jacm/KhamisNPSW24},
while soft unification replaces symbolic matching with similarity over
learned
embeddings~\cite{DBLP:conf/nips/MaeneR23,DBLP:conf/nips/Rocktaschel017}.
Other extensions add temporal operators~\cite{huang2025laser} or expose
foundation models as relations~\cite{li2024vieira}. Each extension
involves different trade-offs, which we align
 with application needs.

\paragraph{Contributions.}
We make the following contributions:
\begin{itemize}
\item A taxonomy of rule-based languages used in
neurosymbolic AI (§\ref{sec:taxonomy});
\item An analysis of recent formalism usage across four research
communities, based on over 50 neurosymbolic systems and applications,
most from the past six years (§\ref{sec:domains}); 
\item A decision matrix mapping application scenarios to required 
features (§\ref{sec:cross});
\item A roadmap of open problems
(§\ref{sec:roadmap}).
\end{itemize}

\section{Logic-Based Languages and Their Semantics}
\label{sec:background}

\paragraph{Classical Datalog.}
Datalog is a function-free language of definite Horn clauses evaluated
under least-fixpoint semantics~\cite{DBLP:books/aw/AbiteboulHV95,DBLP:journals/tkde/CeriGT89}. A program
is a finite set of rules $h \leftarrow b_1, \dots, b_n$ over a finite
relational schema and database facts are rules with empty
bodies. Bottom-up evaluation iteratively applies the immediate consequence operator $T_P$
up to a fixpoint,
\begin{equation}
T_P(I) = \bigl\{\, h\sigma \bigm| (h \leftarrow b_1, \dots, b_n) \in P,\;
b_i\sigma \in I \text{ for all } i \,\bigr\},
\quad
M_P = \mathrm{lfp}(T_P),
\label{eq:tp}
\end{equation}
where $\sigma$ ranges over substitutions grounding all variables of the
rule. Since $T_P$ is monotone over the powerset of the finite Herbrand
base, its least fixpoint exists by the Knaster--Tarski theorem and the
bottom-up iteration reaches it in finitely many steps. The safety
condition requires every head variable to occur in a positive body
atom, so each rule application produces ground facts. Modern engines
(e.g., \system{Nemo}~\cite{DBLP:conf/kr/Ivliev0MSK24}, \system{Souffl\'e}~\cite{DBLP:conf/cav/JordanSS16},  \system{RDFox}~\cite{DBLP:conf/semweb/NenovPMHWB15}) scale this to billions of facts~\cite{DBLP:conf/rweb/Krotzsch25}.

\paragraph{Datalog with negation.}
Negation in rule bodies is interpreted as \emph{negation as failure}
(NAF): $\lnot p$ holds when $p$ cannot be derived. NAF makes the $T_P$
operator non-monotone in general, so a least model need not exist.
\emph{Stratified negation} partitions rules so that negated
predicates are defined in strictly lower strata, producing a canonical
minimal model, the perfect model, that is independent of the chosen
stratification~\cite{DBLP:books/aw/AbiteboulHV95}. The \emph{well-founded
semantics}~\cite{DBLP:journals/jacm/GelderRS91} assigns true/false/unknown to every ground
atom, handling non-stratifiable programs. \emph{Stable model
semantics}~\cite{DBLP:conf/iclp/GelfondL88} underlies ASP
and enables non-monotonic reasoning, as exploited by
\system{dPASP}~\cite{geh2024dpasp}, \system{SLASH}~\cite{skryagin2022slash}, and \system{NeurASP}~\cite{DBLP:conf/ijcai/YangIL20}.
NAF thus checks derivability, whereas classical negation simply flips
the truth value of $p$, as in the fuzzy dialects below.

\paragraph{Aggregation.}
Neurosymbolic tasks often summarise derived relations, e.g., counting
objects in a scene or aggregating probabilistic outputs, which requires
operators such as count, sum, and
min/max~\cite{DBLP:conf/pods/RossS92}. The challenge is their interaction with
recursion. Restricting to aggregates monotone under a suitable order on their
values preserves least-fixpoint evaluation.

\paragraph{Probabilistic logic programming.}
Further extensions attach probabilities to facts or rules.
\system{ProbLog}~\cite{DBLP:conf/ijcai/RaedtKT07} instantiates the distribution semantics of
Sato~\cite{DBLP:conf/iclp/Sato95}. Each probabilistic fact $f$ in a finite set $F$ is
an independent Bernoulli choice with probability $p_f$, and each total
choice $\omega \subseteq F$ determines a possible world, the least
model of $\omega$ together with the rules $P$ of the program. The
probability of a query $q$ sums the weights of the worlds that entail it,
\begin{equation}
\Pr(q) \;=\;
\sum_{\substack{\omega \subseteq F \\ P \,\cup\, \omega \,\models\, q}}
\;\prod_{f \in \omega} p_f
\prod_{f \in F \setminus \omega} \bigl(1 - p_f\bigr),
\label{eq:wmc}
\end{equation}
and is computed in practice by weighted model counting (WMC) over a
propositional encoding of the program~\cite{DBLP:journals/tplp/FierensBRSGTJR15}.
\system{DeepProbLog}~\cite{DBLP:journals/ai/ManhaeveDKDR21} extends this with \emph{neural predicates},
implemented as neural annotated disjunctions whose probabilities are the
output distribution of a network trained end-to-end through the WMC objective. Generative Datalog~\cite{DBLP:conf/icdt/BaranyCKOV16} adds sampling
from discrete distributions, defining a distribution over instances
that is independent of execution order.

\paragraph{Provenance semirings: a unifying algebra.}
Green, Karvounarakis, and Tannen~\cite{DBLP:conf/pods/GreenKT07} showed
that the evaluation of positive relational algebra can be parameterised
by any commutative semiring $(K, \oplus, \otimes, 0, 1)$, and that the
construction extends to recursive Datalog when $K$ is
$\omega$-continuous. An output tuple's annotation is a sum over its
derivation trees, where each tree contributes the product of the
annotations of the input facts at its leaves, counted with multiplicity:
\begin{equation}
\mathrm{ann}(t) \;=\;
\bigoplus_{\tau \in \mathrm{Trees}(t)}\;
\bigotimes_{\ell \in \mathrm{leaves}(\tau)} \mathrm{ann}(\ell),
\label{eq:prov}
\end{equation}
where $\mathrm{leaves}(\tau)$ denotes the multiset of leaf occurrences
in $\tau$. For recursive programs the set $\mathrm{Trees}(t)$ may be
infinite, so $\omega$-continuity ensures that the sum in
Eq.~\ref{eq:prov} is well-defined, with convergence over general
(pre-)semirings characterised
in~\cite{DBLP:journals/jacm/KhamisNPSW24}. Instantiating $K$ with the
Boolean semiring recovers set semantics, and with $\mathbb{N}$ bag
semantics~\cite{DBLP:conf/pods/GreenKT07,DBLP:journals/jacm/KhamisNPSW24}.
Possible-world probabilities, by contrast, do not arise from a direct
semiring homomorphism (derivations may share facts, breaking
independence), but from weighted model counting over the Boolean
provenance
polynomial~\cite{DBLP:conf/pods/GreenKT07,DBLP:journals/tplp/FierensBRSGTJR15}.
This algebra is the foundation on which the differentiable engines
below parameterise their evaluation.

\paragraph{Differentiable evaluation.} To enable end-to-end learning, gradients
must flow from a symbolic loss back to network parameters. Scallop~\cite{DBLP:conf/nips/HuangLCSNSS21,DBLP:journals/pacmpl/LiHN23} achieves this
with a differentiable provenance semiring, computing annotations in the forward
pass and propagating gradients in the backward pass. Exact WMC is intractable,
so in practice it runs approximate semirings (§\ref{sec:axis4}).

\paragraph{Soft unification.}
Classical unification matches two constants if and only if they are the
same symbol, whereas in knowledge graphs and text distinct symbols can
denote the same entity~\cite{DBLP:journals/pvldb/GetoorM12}. The
\system{Neural Theorem Prover}~\cite{DBLP:conf/nips/Rocktaschel017}
relaxes symbolic matching to a differentiable similarity over learned
embeddings, used as a proof score. \system{DeepSoftLog}~\cite{DBLP:conf/nips/MaeneR23} gives
soft unification a probabilistic semantics,
\begin{equation}
\Pr(a \approx b) \;=\; \exp\bigl(-d(e_a, e_b)\bigr),
\label{eq:softunif}
\end{equation}
where $e_a$ and $e_b$ are the learned embeddings of $a$ and $b$ and $d$
is a metric on the embedding space, so the value lies in $(0,1]$ and
$\Pr(a \approx a) = 1$. Treating these pairwise probabilities as
independent facts, however, admits worlds in which $\approx$ is not
transitive. The equivalence semantics of Maene and
Tsamoura~\cite{maene2025equivalence} repairs this by placing a distribution over
equivalence relations instead.

\paragraph{Fuzzy and t-norm semantics.}
Fuzzy dialects assign truth values in $[0,1]$ and compose them with a
t-norm $T$ for conjunction (e.g., the product t-norm $T(x,y) = xy$ or
the G\"odel t-norm $\min(x,y)$) and a t-conorm for disjunction. \system{Logic Tensor Networks}~\cite{badreddine2022ltn},
\system{Logical Neural Networks} (\system{LNN}s)~\cite{sen2022lnn}, and \system{Lifted Relational
Neural Networks}~\cite{sourek2018lrnn} evaluate programs over fuzzy
interpretations rather than distributions over worlds. \system{LNN}s additionally track lower and upper bounds on each truth value,
but no dialect in this family defines a probability distribution over
possible worlds, so degrees of truth and belief are conflated.
The choice also affects trainability. The probabilistic (WMC) objective
is multilinear in the per-atom marginals and depends only on the models
of the constraint, not on its syntactic form. T-norm relaxations are
instead syntax-sensitive, as logically equivalent formulas can induce
different losses, and several operator families suffer from vanishing or
single-argument gradients~\cite{DBLP:journals/ai/KriekenAH22}. Semantic
loss~\cite{DBLP:conf/icml/XuZFLB18} was introduced to restore syntax independence
by computing the constraint probability exactly.

\section{A Taxonomy of Rule-Based Neurosymbolic Languages}
\label{sec:taxonomy}

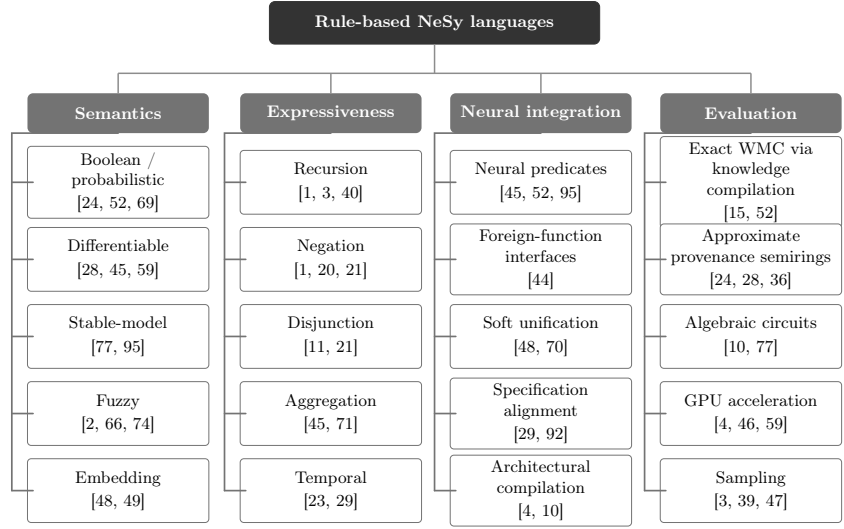
\begin{figure}[t]
\centering
\resizebox{0.9\textwidth}{!}{%
\begin{tikzpicture}[font=\small, >=Latex,
  root/.style ={draw=black!80, fill=black!80, text=white, rounded corners=3pt,
                font=\bfseries, align=center, text width=60mm, minimum height=8mm, inner sep=3pt},
  head/.style ={draw=none, fill=black!55, text=white, rounded corners=3pt, font=\bfseries,
                align=center, text width=32mm, minimum height=7.5mm, inner sep=3pt},
  card/.style ={draw=black!55, fill=white, rounded corners=2pt, align=center,
                text width=32mm, minimum height=12mm, inner sep=3pt, font=\footnotesize, line width=0.4pt},
  spine/.style={line width=0.9pt, line cap=round, draw=black!55},
  bus/.style  ={line width=0.8pt, draw=black!45},
  cit/.style  ={scriptsize, text=black!55}]

\node[root] (root) at (5.95,1.7) {Rule-based NeSy languages};
\node[head] (h1) at (0,0)     {Semantics};
\node[head] (h2) at (4.0,0)   {Expressiveness};
\node[head] (h3) at (7.95,0)  {Neural integration};
\node[head] (h4) at (11.9,0)  {Evaluation};
\draw[bus] (5.95,1.3) -- (5.95,0.75);
\draw[bus] (0,0.75) -- (11.9,0.75);
\foreach \x in {0,4.0,7.95,11.9} \draw[bus] (\x,0.75) -- (\x,0.375);

\node[card] (a1) at (0,-1.30) {Boolean / probabilistic\\[1.5pt]{\cite{DBLP:conf/pods/GreenKT07,DBLP:journals/ai/ManhaeveDKDR21,DBLP:conf/ijcai/RaedtKT07}}};
\node[card] (a2) at (0,-2.75) {Differentiable\\[1.5pt]{\cite{DBLP:conf/nips/HuangLCSNSS21,DBLP:journals/pacmpl/LiHN23,naik2025dolphin}}};
\node[card] (a3) at (0,-4.20) {Stable-model\\[1.5pt]{\cite{skryagin2022slash,DBLP:conf/ijcai/YangIL20}}};
\node[card] (a4) at (0,-5.65) {Fuzzy\\[1.5pt]{\cite{badreddine2022ltn,pryor2023neupsl,sen2022lnn}}};
\node[card] (a5) at (0,-7.10) {Embedding\\[1.5pt]{\cite{DBLP:conf/nips/MaeneR23,maene2025equivalence}}};
\draw[spine] (0,-0.375) -- (-2.0,-0.375) -- (-2.0,-7.10);
\foreach \k in {1,...,5} \draw[spine] (-2.0,{0.15-1.45*\k}) -- (-1.6,{0.15-1.45*\k});

\node[card] (b1) at (4.0,-1.30) {Recursion\\[1.5pt]{\cite{DBLP:books/aw/AbiteboulHV95,DBLP:conf/icdt/BaranyCKOV16,DBLP:conf/rweb/Krotzsch25}}};
\node[card] (b2) at (4.0,-2.75) {Negation\\[1.5pt]{\cite{DBLP:books/aw/AbiteboulHV95,DBLP:journals/jacm/GelderRS91,DBLP:conf/iclp/GelfondL88}}};
\node[card] (b3) at (4.0,-4.20) {Disjunction\\[1.5pt]{\cite{DBLP:journals/tods/EiterGM97,DBLP:conf/iclp/GelfondL88}}};
\node[card] (b4) at (4.0,-5.65) {Aggregation\\[1.5pt]{\cite{DBLP:journals/pacmpl/LiHN23,DBLP:conf/pods/RossS92}}};
\node[card] (b5) at (4.0,-7.10) {Temporal\\[1.5pt]{\cite{DBLP:conf/ijcai/GiacomoV13,huang2025laser}}};
\draw[spine] (4.0,-0.375) -- (2.0,-0.375) -- (2.0,-7.10);
\foreach \k in {1,...,5} \draw[spine] (2.0,{0.15-1.45*\k}) -- (2.4,{0.15-1.45*\k});

\node[card] (c1) at (7.95,-1.30) {Neural predicates\\[1.5pt]{\cite{DBLP:journals/pacmpl/LiHN23,DBLP:journals/ai/ManhaeveDKDR21,DBLP:conf/ijcai/YangIL20}}};
\node[card] (c2) at (7.95,-2.75) {Foreign-function\\ interfaces\\[1.5pt]{\cite{li2024vieira}}};
\node[card] (c3) at (7.95,-4.20) {Soft unification\\[1.5pt]{\cite{DBLP:conf/nips/MaeneR23,DBLP:conf/nips/Rocktaschel017}}};
\node[card] (c4) at (7.95,-5.65) {Specification alignment\\[1.5pt]{\cite{huang2025laser,DBLP:conf/icml/XuZFLB18}}};
\node[card] (c5) at (7.95,-7.10) {Architectural\\ compilation\\[1.5pt]{\cite{DBLP:conf/asplos/Biberstein0DN26,derkinderen2025deeplog}}};
\draw[spine] (7.95,-0.375) -- (5.95,-0.375) -- (5.95,-7.10);
\foreach \k in {1,...,5} \draw[spine] (5.95,{0.15-1.45*\k}) -- (6.35,{0.15-1.45*\k});

\node[card] (d1) at (11.9,-1.30) {Exact WMC via\\knowledge compilation\\[1.5pt]{\cite{DBLP:journals/tplp/FierensBRSGTJR15,DBLP:journals/ai/ManhaeveDKDR21}}};
\node[card] (d2) at (11.9,-2.75) {Approximate\\ provenance semirings\\[1.5pt]{\cite{DBLP:conf/pods/GreenKT07,DBLP:conf/nips/HuangLCSNSS21,DBLP:journals/jacm/KhamisNPSW24}}};
\node[card] (d3) at (11.9,-4.20) {Algebraic circuits\\[1.5pt]{\cite{derkinderen2025deeplog,skryagin2022slash}}};
\node[card] (d4) at (11.9,-5.65) {GPU acceleration\\[1.5pt]{\cite{DBLP:conf/asplos/Biberstein0DN26,DBLP:conf/iclr/MaeneDM25,naik2025dolphin}}};
\node[card] (d5) at (11.9,-7.10) {Sampling\\[1.5pt]{\cite{DBLP:conf/icdt/BaranyCKOV16,vankrieken2023anesi,DBLP:conf/icml/MaeneDR24}}};
\draw[spine] (11.9,-0.375) -- (9.9,-0.375) -- (9.9,-7.10);
\foreach \k in {1,...,5} \draw[spine] (9.9,{0.15-1.45*\k}) -- (10.3,{0.15-1.45*\k});
\end{tikzpicture}}
\caption{The four-axis taxonomy of rule-based neurosymbolic languages.}
\label{fig:taxonomy}
\end{figure}

We categorise rule-based neurosymbolic languages along four 
axes (Fig.~\ref{fig:taxonomy}). The axes make explicit the choices involved in selecting a dialect for a given task.

\subsection{Axis 1: Semantics}
The first axis concerns how truth values, probabilities, or belief
degrees are assigned and propagated (Table~\ref{tab:semantics}).
\emph{Boolean-probabilistic} dialects evaluate under possible-world semantics with WMC (Eq.~\ref{eq:wmc}). \system{DeepProbLog} is the canonical
example.\footnote{\system{ProbLog} programs admit function symbols. We
place \system{DeepProbLog} in Fig.~\ref{fig:venn}a by the semantics of
its function-free core, whose distribution semantics requires each
total choice to induce a two-valued well-founded model.} \system{Scallop} recovers the same semantics as one instantiation of its provenance framework, although in practice it runs approximate configurations. The limitation is
scalability (§\ref{sec:axis4}).
\emph{Differentiable-probabilistic} dialects relax exactness for
gradient flow, and \system{Scallop}'s top-$k$ semiring is the standard choice.
\emph{Stable-model} dialects (\system{NeurASP}, \system{SLASH},
\system{dPASP}) place the distribution over answer sets rather than
over least models, importing the non-monotonic reasoning of ASP into
the neural pipeline at a higher inference cost. \emph{Soft-unification} dialects replace symbolic identity
with embedding similarity (Eq.~\ref{eq:softunif}), interpreted under probabilistic
(\system{DeepSoftLog}) or equivalence semantics.

\emph{Fuzzy/t-norm}
dialects operate on a single $[0,1]$ interpretation
(\system{LTN}, \system{LNN}). \emph{Algebraic-circuit} dialects, notably
\system{DeepLog}~\cite{derkinderen2025deeplog,manhaeve2026deeplogsoftware}, make the algebraic
structure itself a \emph{parameter}. By instantiating the algebra appropriately, \system{DeepLog}'s compiled circuits emulate \system{DeepProbLog}, \system{LTN}, or \system{NeurASP}. This generality is achieved at the circuit level rather than at the
language level, so \system{DeepLog} serves as a compilation target rather than
as a source language.

\begin{table}[t]
\centering
\caption{Axis 1: semantic foundation of rule-based neurosymbolic dialects.}
\label{tab:semantics}
\begin{tabular}{@{}lll@{}}
\toprule
\textbf{Category} & \textbf{Semantics} & \textbf{Representative system(s)}\\
\midrule
Boolean-probabilistic & possible-world distribution (WMC) & DeepProbLog, Scallop\\
Differentiable-prob. & approx.\ WMC / fuzzy provenance & Scallop, DOLPHIN\\
Stable-model (ASP) & distribution over answer sets & NeurASP, SLASH, dPASP\\
Soft-unif./embedding & probabilistic embedding equality & DeepSoftLog\\
Fuzzy / t-norm & $[0,1]$ degrees, no world distribution & LTN, LNN, NeuPSL\\
Algebraic-circuit & algebra as circuit parameter & DeepLog\\
\bottomrule
\end{tabular}
\end{table}

\subsection{Axis 2: Expressiveness}
\begin{figure}[t]
\centering
\begin{subfigure}[b]{0.49\textwidth}\centering
\resizebox{\linewidth}{!}{%
\begin{tikzpicture}[
  ring/.style ={thick, draw=black!60},
  rname/.style={font=\large\bfseries, align=center},
  sys/.style  ={font=\large, text=black!70, align=center}]
\draw[ring, fill=black!22] (0, 1.40)  ellipse (5.5  and 6.00);
\draw[ring, fill=black!17] (0, 0.875) ellipse (4.85 and 5.125);
\draw[ring, fill=black!12] (0, 0.05)  ellipse (4.35 and 3.95);
\draw[ring, fill=black!7]  (0,-0.775) ellipse (3.40 and 2.775);
\draw[ring, fill=black!3]  (0,-1.60)  ellipse (2.40 and 1.60);
\node[rname] at (0,6.65) {Disjunctive\\Datalog};
\node[rname] at (0,5.25) {Datalog with\\Stable Negation};
\node[sys]   at (0,4.40) {NeurASP, SLASH, dPASP};
\node[rname] at (0,3.25) {Well-founded\\Datalog};
\node[sys]   at (0,2.40) {DeepProbLog};
\node[rname] at (0,1.25) {Stratified\\Datalog};
\node[sys]   at (0,0.40) {Scallop, Lobster, DOLPHIN};
\node[rname] at (0,-0.80) {Datalog};
\node[sys]   at (0,-1.50) {TensorLog (chain-rule)};
\node[sys]   at (0,-2.15) {Scallop\,v0};
\end{tikzpicture}}
\caption{Datalog-based dialects for NeSy.}
\label{fig:venn-a}
\end{subfigure}\hfill
\begin{subfigure}[b]{0.49\textwidth}\centering
\resizebox{\linewidth}{!}{%
\begin{tikzpicture}[
  reg/.style    ={thick, draw=black!60},
  apart/.style  ={thick, draw=black!60, dash pattern=on 3pt off 2pt},
  rname/.style  ={font=\large\bfseries, align=center},
  grp/.style    ={font=\large\bfseries, text=black!75, align=center},
  sys/.style    ={font=\large, text=black!70, align=center}]
\draw[reg, fill=black!3] (0,0.1) ellipse (5.6 and 6.0);
\node[grp] at (0,5.05) {Non-Datalog formalisms};
\draw[reg, fill=black!9] (0,1.45) ellipse (4.6 and 3.0);
\node[rname] at (0,3.95) {Weighted FOL};
\draw[reg, fill=black!17] (0,2.50) ellipse (3.35 and 1.05);
\node[rname] at (0,2.85) {Fuzzy FOL};
\node[sys]   at (0,2.15) {LTN, LNN, NeuPSL};
\draw[reg, fill=black!17] (0,0.10) ellipse (3.35 and 1.05);
\node[rname] at (0,0.45) {Markov logic};
\node[sys]   at (0,-0.25) {pLogicNet, ExpressGNN};
\draw[reg, fill=black!13] (-2.55,-2.90) ellipse (2.15 and 1.15);
\node[rname] at (-2.55,-2.55) {Functional DSL};
\node[sys]   at (-2.55,-3.25) {NS-CL, LEFT};
\draw[reg, fill=black!13] (2.55,-2.90) ellipse (2.15 and 1.15);
\node[rname] at (2.55,-2.55) {Temporal\\(LTL/STL)};
\node[sys]   at (2.55,-3.40) {LASER};
\draw[apart, fill=black!13] (0,-4.75) ellipse (2.6 and 0.95);
\node[rname] at (0,-4.45) {Propositional SAT};
\node[sys]   at (0,-5.05) {SATNet};
\end{tikzpicture}}
\caption{Other formalisms for NeSy.}
\label{fig:venn-b}
\end{subfigure}
\caption{Rule-based languages for neurosymbolic (NeSy) AI.}
\label{fig:venn}
\end{figure}
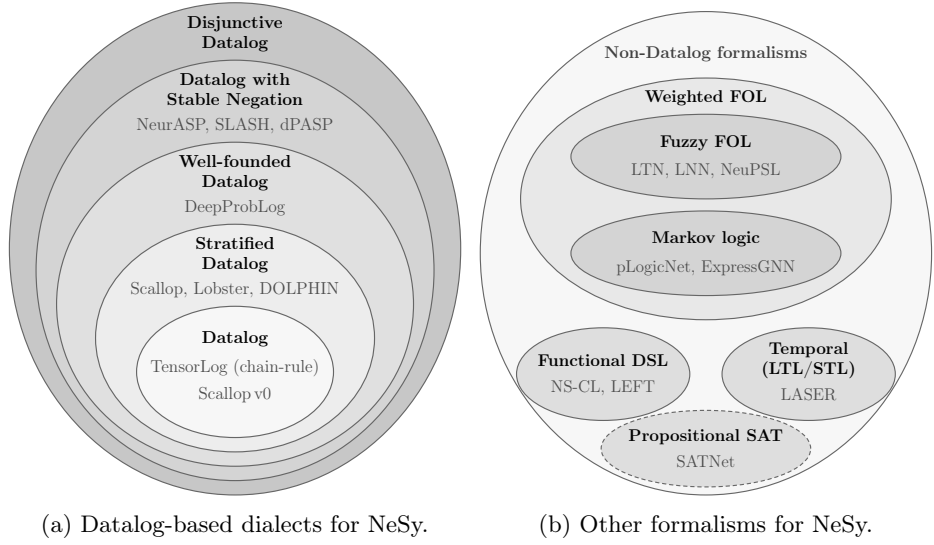

The second axis concerns which syntactic features a dialect supports,
since these determine the reasoning patterns it can express.
Figure~\ref{fig:venn} captures the main dialects, while Table~\ref{tab:formalism}
classifies formalism families and their features.

The Venn diagram (Fig.~\ref{fig:venn}a) illustrates the inclusions among these dialects, so a task can be placed in the innermost ring that supports it.
Negation moves a task from the positive to the stratified ring, recursion through negation to the well-founded one, non-monotonicity to the stable one, and head disjunction to the outermost ring.
The inner inclusions are
syntactic, whereas the well-founded and stable rings distinguish
alternative \emph{semantics} for normal programs, which coincide on
stratified
ones~\cite{DBLP:journals/jacm/GelderRS91,DBLP:conf/iclp/GelfondL88}.
The probabilistic semantics attached to each ring (top-$k$, possible
worlds, weighted stable models) are not themselves ordered, so a system
can inherit a sublanguage's expressiveness without inheriting a
comparable inference cost. Fig.~\ref{fig:venn}b collects families outside any such inclusion order. Fuzzy logic and Markov logic share only a weighted first-order basis. Propositional SAT, though expressively weaker, is kept separate, as its differentiable solver forms a distinct computational paradigm rather than a sublanguage.

\emph{Positive Datalog} is expressive enough
for digit recognition and simple planning in the original
\system{Scallop} prototype~\cite{DBLP:conf/nips/HuangLCSNSS21} but cannot encode an interpreter for \system{CLEVR}'s functional
query language~\cite{DBLP:conf/cvpr/JohnsonHMFZG17,DBLP:journals/pacmpl/LiHN23}.
\emph{Stratified
negation and aggregation} lift this limitation~\cite{DBLP:journals/pacmpl/LiHN23}.
\emph{Foreign-function interfaces (FFI)} let programs invoke external computations,
including foundation-model calls, as relational
predicates~\cite{li2024vieira}. Such calls are non-deterministic and
potentially stateful, which the \system{Vieira} system controls by treating
models as \emph{stateless} relations (§\ref{sec:axis3}). \emph{Temporal} extensions add operators over time:
\system{LASER}'s spatio-temporal specification language
(STSL)~\cite{huang2025laser} builds on finite-trace linear
temporal logic (LTL$_f$)~\cite{DBLP:conf/ijcai/GiacomoV13}. 
\emph{Recursive probabilistic} programs,
formalised in Generative Datalog~\cite{DBLP:conf/icdt/BaranyCKOV16},
subsume both deterministic recursion and probabilistic generation, the
most expressive combination on the probabilistic side of the hierarchy.
Differentiable \emph{rule learning} (ILP) lies outside this hierarchy:
$\partial$\system{ILP}~\cite{evans2018dilp}, $\alpha$\system{ILP}~\cite{shindo2023alphailp},
and \system{DiffLog}~\cite{si2019difflog} learn Boolean rules via a
$[0,1]$ relaxation used only during training.

\begin{table}[t]
\centering
\caption{Axis~2: formalism families and the features they support
(above the rule: the Datalog/ASP hierarchy of Fig.~\ref{fig:venn}a;
below it: formalisms outside it, Fig.~\ref{fig:venn}b).}
\label{tab:formalism}
\footnotesize
\setlength{\tabcolsep}{2pt}
\renewcommand{\arraystretch}{1.05}
\begin{tabular}{@{}llccc@{\hspace{2pt}}c@{}}
\toprule
\textbf{Family} & \textbf{Systems} &
\textbf{Rec.} & \textbf{Neg.} & \textbf{Agg.} & \textbf{Uncert.}\\
\midrule
Datalog (positive)      & Scallop\,v0                         & \cmark & --      & --      & top-$k$\\
Datalog (chain-rule)    & TensorLog, NeuralLP, DRUM           & bnd    & --      & partial & soft\\
Datalog (neg{+}agg)     & Scallop, Lobster, DOLPHIN           & \cmark & strat.  & \cmark  & semiring\\
ProbLog ({+}neural)     & DeepProbLog                         & \cmark & NAF     & --      & worlds\\
Prolog (SDCG/soft)      & DeepStochLog, DeepSoftLog           & \cmark & --      & --      & der./worlds\\
ASP (prob./circuits)    & NeurASP, SLASH, dPASP               & \cmark & n.m.    & \cmark  & multiple\\
ILP (differentiable)    & $\partial$ILP, $\alpha$ILP, DiffLog & bnd    & --      & --      & relax.\\
\midrule
Fuzzy FOL               & LTN, LNN, NeuPSL                    & --     & class.  & --      & $[0,1]$\\
Markov logic            & pLogicNet, ExpressGNN               & --     & --      & --      & MLN\\
Functional DSL          & NS-CL, LEFT                         & --     & partial & partial & soft\\
Temporal (LTL/STL)      & LASER                               & n.a.   & n.a.    & n.a.    & quant.\\
Propositional SAT       & SATNet                              & --     & --      & --      & relax.\\
\bottomrule
\end{tabular}\par\smallskip
{\scriptsize\raggedright
bnd = bounded; strat.\ = stratified; n.m.\ = non-monotonic;
class.\ = classical; der.\ = derivations;
quant.\ = quantitative satisfaction; relax.\ = continuous relaxation;
MLN = Markov-logic weights; SDCG = stochastic definite clause
grammars.\par}
\end{table}

\subsection{Axis 3: Neural Integration}\label{sec:axis3}
The third axis describes how the rule layer interfaces with neural
components. \emph{Neural predicates} inject a network's output
distribution as probabilistic facts, and the network is trained
end-to-end by differentiating through
evaluation~\cite{DBLP:journals/pacmpl/LiHN23,DBLP:journals/ai/ManhaeveDKDR21}.
\emph{Foreign-function interfaces} instead treat foundation models as stateless
relational functions, invoked with zero or few in-context examples and
never fine-tuned~\cite{li2024vieira}. \emph{Soft unification} modifies the inference mechanism itself, associating each
constant with a learned
embedding~\cite{DBLP:conf/nips/MaeneR23,DBLP:conf/nips/Rocktaschel017}.

\begin{figure}[t]
\centering
\begin{tikzpicture}[>=Latex,
  stage/.style={draw=black, line width=0.6pt, rounded corners=2.5pt, fill=white,
                align=center, font=\scriptsize,
                text width=2.08cm, minimum height=1.25cm, inner sep=2pt},
  lbl/.style  ={font=\scriptsize, inner sep=0pt, above=3pt},
  flow/.style ={line width=0.6pt, -{Latex[length=2.2mm]}}]
\node[stage] (target) at (0,0)
  {Target output\\ $\vec y$ (e.g.\ sum $=2$)};
\node[stage] (proofs) at (3.31,0)
  {Symbolic proofs\\ entailing $\vec y$ under $P$};
\node[stage] (diff) at (6.62,0)
  {Differentiable function\\ WMC, top-$k$, fuzzy relaxation};
\node[stage] (neural) at (9.93,0)
  {Neural \\component\\ gradient update of $\neural$};
\draw[flow] (target) -- node[lbl]{abduce}  (proofs);
\draw[flow] (proofs) -- node[lbl]{compile} (diff);
\draw[flow] (diff)   -- node[lbl]{loss, $\nabla$} (neural);
\end{tikzpicture}
\caption{Abduction-based neural integration~\cite{neurolog,wang2023multiinstance}.
Proofs of the target are abduced from the program and compiled into a
differentiable training loss.}
\label{fig:abduction}
\end{figure}
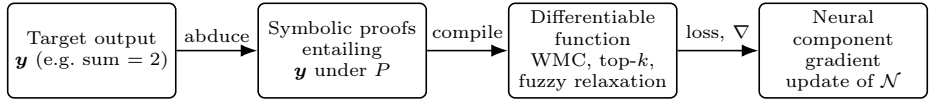

\emph{Abduction} runs the interface in the reverse direction
(Fig.~\ref{fig:abduction}). From a target output, the symbolic
component abduces the proofs, i.e., the combinations of hidden labels
that entail the target. Compiling these proofs into a differentiable
function, e.g., a WMC circuit, a top-$k$ provenance semiring, or a
fuzzy relaxation, produces the loss that trains the network.
\system{NeuroLog} constructs this coupling
compositionally~\cite{neurolog}, and Wang et
al.~\cite{wang2023multiinstance} characterise the weak-supervision
settings under which such learning is possible.

\emph{Specification alignment} uses the program as a supervision
signal. In \system{LASER}, an LLM extracts temporal specifications and
a differentiable checker scores predicted scene graphs against
them~\cite{huang2025laser}. Finally, \emph{architecture-level
integration} compiles programs to low-level execution targets.
\system{Lobster}~\cite{DBLP:conf/asplos/Biberstein0DN26} emits GPU kernels
(§\ref{sec:axis4}), while
\system{DeepLog}~\cite{derkinderen2025deeplog} compiles to algebraic
circuits.

\paragraph{Beyond the explicit rule layer.}
The integration mechanisms discussed so far expose an inspectable
Datalog/ASP rule layer. A large, complementary family instead pairs the network with a lifted graphical model or a fuzzy loss, a composite design surveyed by Feldstein et al.~\cite{DBLP:journals/corr/abs-2410-22077}. In these systems, the
symbolic side \emph{regularises} training rather than carrying the
reasoning itself. Two schemes recur (Fig.~\ref{fig:composite-direct}).
\emph{Parallel} supervision penalises the divergence between the
distributions $\probneural, \problogical$ induced by the neural and
logical components $\neural$ and $\logical$, which presupposes a
logical theory blended with an uncertainty formalism such as
probability theory or fuzzy sets. \emph{Stratified} supervision, by contrast, treats the predictions of
the neural component as atoms and adds a satisfiability term over them. 
Semantic loss~\cite{DBLP:conf/icml/XuZFLB18} instantiates this scheme
and collapses to standard weighted model counting
once the constraint is compiled rather than
approximated. The composite architectures of Fig.~\ref{fig:composite-parallel}
(Teacher--Student, Deep Probabilistic Logic, Concordia)
build on Markov logic or probabilistic soft logic (PSL), the
weighted-FOL region of Fig.~\ref{fig:venn}b, and thus provide no auditable rule
layer. Teacher--Student and DPL
use $\logical$ only during training and return
$\vec{\hat y}_{\neural}$, whereas Concordia retains $\logical$ at inference, using a learned gate
$\gating$ to return a mixture $\vec{\hat y}_{\concordia}$.

\begin{figure}[t!]
\centering
\captionsetup[subfigure]{aboveskip=2pt, belowskip=0pt}

\begin{subfigure}[b]{\linewidth}
\centering
\begin{tikzpicture}
	\begin{pgfonlayer}{nodelayer}
		\node [style=none] (0) at (0, 0.25) {};
		\node [style=none] (1) at (0, 1) {};
		\node [style=none] (2) at (0.5, 1) {};
		\node [style=none] (3) at (0.5, 0.5) {};
		\node [style=none] (4) at (0.5, 1.5) {};
		\node [style=none] (5) at (2, 1.5) {};
		\node [style=none] (6) at (2, 0.5) {};
		\node [style=none] (11) at (0, 0) {$\inputcolour{\vec{x}}$};
		\node [style=none] (12) at (0, -0.25) {};
		\node [style=none] (13) at (0, -1) {};
		\node [style=none] (14) at (0.5, -1) {};
		\node [style=none] (15) at (0.5, -0.5) {};
		\node [style=none] (16) at (0.5, -1.5) {};
		\node [style=none] (17) at (2, -1.5) {};
		\node [style=none] (18) at (2, -0.5) {};
		\node [style=none] (19) at (2.75, -1) {};
		\node [style=none] (21) at (2.75, 1) {};
		\node [style=none] (26) at (2, -1) {};
		\node [style=none] (32) at (1.25, -1) {$\logicalcolour{\logical}$};
		\node [style=none] (33) at (1.25, 1) {$\neuralcolour{\neural}$};
		\node [style=none] (111) at (4, 0) {$\gatingcolour{\vec{\hat{y}}}$};
		\node [style=none] (114) at (3.75, 0) {};
		\node [style=smallgreen] (117) at (2.75, 0) {$\gatingcolour{+}$};
		\node [style=none] (145) at (2, 1) {};
		\node [style=none] (146) at (2.375, 1.25) {$\neuralcolour{\probneural}$};
		\node [style=none] (147) at (2.375, -1.25) {$\logicalcolour{\problogical}$};
		\node [style=none] (148) at (6, -0.5) {};
		\node [style=none] (149) at (6, 0.5) {};
		\node [style=none] (150) at (7.5, 0.5) {};
		\node [style=none] (151) at (7.5, -0.5) {};
		\node [style=none] (152) at (5, 0) {$\inputcolour{\vec{x}}$};
		\node [style=none] (153) at (5.25, 0) {};
		\node [style=none] (154) at (6, 0) {};
		\node [style=none] (155) at (8.25, 0.5) {};
		\node [style=none] (156) at (8.25, -0.5) {};
		\node [style=none] (157) at (9.75, -0.5) {};
		\node [style=none] (158) at (9.75, 0.5) {};
		\node [style=none] (159) at (9.75, 0) {};
		\node [style=none] (160) at (10.5, 0) {};
		\node [style=none] (161) at (7.5, 0) {};
		\node [style=none] (162) at (9.75, 0) {};
		\node [style=none] (163) at (9, 0) {$\logicalcolour{\logical}$};
		\node [style=none] (164) at (6.75, 0) {$\neuralcolour{\neural}$};
		\node [style=none] (165) at (10.75, 0) {$\neuralcolour{\vec{\hat{y}}}$};
	\end{pgfonlayer}
	\begin{pgfonlayer}{edgelayer}
		\draw [style=orangesolid] (4.center) to (3.center);
		\draw [style=orangesolid] (3.center) to (6.center);
		\draw [style=orangesolid] (4.center) to (5.center);
		\draw [style=orangesolid] (5.center) to (6.center);
		\draw [style=black] (1.center) to (2.center);
		\draw (0.center) to (1.center);
		\draw (12.center) to (13.center);
		\draw [style=black] (13.center) to (14.center);
		\draw [style=bluesolid] (16.center) to (17.center);
		\draw [style=bluesolid] (18.center) to (17.center);
		\draw [style=bluesolid] (15.center) to (16.center);
		\draw [style=bluesolid] (15.center) to (18.center);
		\draw [style=greendirected] (117) to (114.center);
		\draw [style=orangedirected] (21.center) to (117);
		\draw [style=bluedirected] (19.center) to (117);
		\draw [style=bluesolid] (26.center) to (19.center);
		\draw [style=orangesolid] (145.center) to (21.center);
		\draw [style=orangesolid] (149.center) to (148.center);
		\draw [style=orangesolid] (148.center) to (151.center);
		\draw [style=orangesolid] (149.center) to (150.center);
		\draw [style=orangesolid] (150.center) to (151.center);
		\draw [style=black] (153.center) to (154.center);
		\draw [style=bluesolid] (156.center) to (157.center);
		\draw [style=bluesolid] (158.center) to (157.center);
		\draw [style=bluesolid] (155.center) to (156.center);
		\draw [style=bluesolid] (155.center) to (158.center);
		\draw [style=orangedashed] (161.center) to (162.center);
		\draw [style=orangedirected] (159.center) to (160.center);
	\end{pgfonlayer}
\end{tikzpicture}\caption{Parallel (left) and stratified (right) architectures. In the
parallel scheme, the divergence between $\probneural$ and $\problogical$
regularises the network and $\vec{\hat y}$ composes both predictions.
In the stratified
scheme, a satisfiability term over the neural predictions penalises
violations, the construction behind the semantic loss~\cite{DBLP:conf/icml/XuZFLB18}.}
\label{fig:composite-direct}
\end{subfigure}

\vspace{4ex}

\begin{subfigure}[b]{\linewidth}
\centering
\begin{tikzpicture}
	\begin{pgfonlayer}{nodelayer}
		\node [style=none] (0) at (0, 0.25) {};
		\node [style=none] (1) at (0, 1) {};
		\node [style=none] (2) at (0.5, 1) {};
		\node [style=none] (3) at (0.5, 0.5) {};
		\node [style=none] (4) at (0.5, 1.5) {};
		\node [style=none] (5) at (2, 1.5) {};
		\node [style=none] (6) at (2, 0.5) {};
		\node [style=none] (7) at (2, 1) {};
		\node [style=none] (8) at (2.5, 1) {};
		\node [style=none] (11) at (0, 0) {$\inputcolour{\vec{x}}$};
		\node [style=none] (12) at (0, -0.25) {};
		\node [style=none] (13) at (0, -1) {};
		\node [style=none] (14) at (0.5, -1) {};
		\node [style=none] (15) at (0.5, -0.5) {};
		\node [style=none] (16) at (0.5, -1.5) {};
		\node [style=none] (17) at (2, -1.5) {};
		\node [style=none] (18) at (2, -0.5) {};
		\node [style=none] (19) at (1.25, -0.5) {};
		\node [style=none] (20) at (1.25, -0.125) {};
		\node [style=none] (21) at (1.25, 0.5) {};
		\node [style=none] (22) at (1.25, 0.125) {};
		\node [style=none] (23) at (0.75, 0.25) {$\neuralcolour{\probneural}$};
		\node [style=none] (24) at (0.75, -0.25) {$\logicalcolour{\problogical}$};
		\node [style=none] (26) at (2, -1) {};
		\node [style=none] (32) at (1.25, -1) {$\logicalcolour{\logical}$};
		\node [style=none] (33) at (1.25, 1) {$\neuralcolour{\neural}$};
		\node [style=none] (34) at (0.25, 0) {};
		\node [style=none] (74) at (7.75, 0.25) {};
		\node [style=none] (75) at (7.75, 1) {};
		\node [style=none] (76) at (8.25, 1) {};
		\node [style=none] (77) at (8.25, 0.5) {};
		\node [style=none] (78) at (8.25, 1.5) {};
		\node [style=none] (79) at (9.75, 1.5) {};
		\node [style=none] (80) at (9.75, 0.5) {};
		\node [style=none] (81) at (9.75, 1) {};
		\node [style=none] (82) at (10.5, 1) {};
		\node [style=none] (83) at (11, 1.5) {};
		\node [style=none] (84) at (10.5, 1.5) {};
		\node [style=none] (85) at (7.75, 0) {$\inputcolour{\vec{x}}$};
		\node [style=none] (86) at (7.75, -0.25) {};
		\node [style=none] (87) at (7.75, -1) {};
		\node [style=none] (88) at (8.25, -1) {};
		\node [style=none] (89) at (8.25, -0.5) {};
		\node [style=none] (90) at (8.25, -1.5) {};
		\node [style=none] (91) at (9.75, -1.5) {};
		\node [style=none] (92) at (9.75, -0.5) {};
		\node [style=none] (93) at (9.25, -0.5) {};
		\node [style=none] (94) at (9.25, 0.5) {};
		\node [style=none] (95) at (8.75, 0.5) {};
		\node [style=none] (96) at (8.75, -0.5) {};
		\node [style=none] (97) at (8.5, 0.25) {$\neuralcolour{\vec{\hat{y}}_\neural}$};
		\node [style=none] (98) at (9.5, -0.25) {$\logicalcolour{\problogical}$};
		\node [style=none] (99) at (10.5, -1) {};
		\node [style=none] (100) at (9.75, -1) {};
		\node [style=none] (101) at (10.5, -1.5) {};
		\node [style=none] (102) at (11, -1.5) {};
		\node [style=none] (103) at (10.75, 0) {$\gatingcolour{\gating}$};
		\node [style=none] (104) at (11, 0) {};
		\node [style=none] (105) at (11.5, 0) {};
		\node [style=none] (106) at (9, -1) {$\logicalcolour{\logical}$};
		\node [style=none] (107) at (9, 1) {$\neuralcolour{\neural}$};
		\node [style=none] (108) at (8, 0) {};
		\node [style=none] (109) at (10.5, 0) {};
		\node [style=none] (110) at (11.75, 0) {$\gatingcolour{\vec{\hat{y}}_\concordia}$};
		\node [style=none] (111) at (1.25, 0) {$\gatingcolour{q}$};
		\node [style=none] (112) at (1.5, 0) {};
		\node [style=none] (113) at (1.75, 0) {};
		\node [style=none] (114) at (1.75, 0.5) {};
		\node [style=none] (115) at (1.75, 0) {};
		\node [style=none] (116) at (2.75, 1) {$\neuralcolour{\vec{\hat{y}}_\neural}$};
		\node [style=none] (117) at (4, 0.25) {};
		\node [style=none] (118) at (4, 1) {};
		\node [style=none] (119) at (4.5, 1) {};
		\node [style=none] (120) at (4.5, 0.5) {};
		\node [style=none] (121) at (4.5, 1.5) {};
		\node [style=none] (122) at (6, 1.5) {};
		\node [style=none] (123) at (6, 0.5) {};
		\node [style=none] (124) at (6, 1) {};
		\node [style=none] (125) at (6.5, 1) {};
		\node [style=none] (126) at (4, 0) {$\inputcolour{\vec{x}}$};
		\node [style=none] (127) at (4, -0.25) {};
		\node [style=none] (128) at (4, -1) {};
		\node [style=none] (129) at (4.5, -1) {};
		\node [style=none] (130) at (4.5, -0.5) {};
		\node [style=none] (131) at (4.5, -1.5) {};
		\node [style=none] (132) at (6, -1.5) {};
		\node [style=none] (133) at (6, -0.5) {};
		\node [style=none] (134) at (5.25, -0.5) {};
		\node [style=none] (135) at (5.25, -0.125) {};
		\node [style=none] (136) at (5.25, 0.5) {};
		\node [style=none] (137) at (5.25, 0.125) {};
		\node [style=none] (138) at (4.75, 0.25) {$\neuralcolour{\probneural}$};
		\node [style=none] (139) at (4.75, -0.25) {$\logicalcolour{\problogical}$};
		\node [style=none] (141) at (5.25, -1) {$\logicalcolour{\logical}$};
		\node [style=none] (142) at (5.25, 1) {$\neuralcolour{\neural}$};
		\node [style=none] (143) at (4.25, 0) {};
		\node [style=none] (144) at (5.25, 0) {$\gatingcolour{q}$};
		\node [style=none] (145) at (5.5, 0) {};
		\node [style=none] (146) at (5.75, 0) {};
		\node [style=none] (147) at (5.75, 0.5) {};
		\node [style=none] (148) at (5.75, 0) {};
		\node [style=none] (149) at (6.75, 1) {$\neuralcolour{\vec{\hat{y}}_\neural}$};
		\node [style=none] (150) at (5.75, -0.5) {};
		\node [style=none] (153) at (10.125, 1.25) {$\neuralcolour{\probneural}$};
		\node [style=none] (154) at (10.125, -1.25) {$\logicalcolour{\problogical}$};
	\end{pgfonlayer}
	\begin{pgfonlayer}{edgelayer}
		\draw [style=orangesolid] (4.center) to (3.center);
		\draw [style=orangesolid] (3.center) to (6.center);
		\draw [style=orangesolid] (4.center) to (5.center);
		\draw [style=orangesolid] (5.center) to (6.center);
		\draw [style=orangedirected] (7.center) to (8.center);
		\draw [style=black] (1.center) to (2.center);
		\draw (0.center) to (1.center);
		\draw (12.center) to (13.center);
		\draw [style=black] (13.center) to (14.center);
		\draw [style=bluesolid] (16.center) to (17.center);
		\draw [style=bluesolid] (18.center) to (17.center);
		\draw [style=bluesolid] (15.center) to (16.center);
		\draw [style=bluesolid] (15.center) to (18.center);
		\draw [style=orangedirected] (21.center) to (22.center);
		\draw [style=bluedirected] (19.center) to (20.center);
		\draw [style=orangesolid] (78.center) to (77.center);
		\draw [style=orangesolid] (77.center) to (80.center);
		\draw [style=orangesolid] (78.center) to (79.center);
		\draw [style=orangesolid] (79.center) to (80.center);
		\draw [style=orangedirected] (81.center) to (82.center);
		\draw [style=black] (75.center) to (76.center);
		\draw (74.center) to (75.center);
		\draw (86.center) to (87.center);
		\draw [style=black] (87.center) to (88.center);
		\draw [style=bluesolid] (90.center) to (91.center);
		\draw [style=bluesolid] (92.center) to (91.center);
		\draw [style=bluesolid] (89.center) to (90.center);
		\draw [style=bluesolid] (89.center) to (92.center);
		\draw [style=orangedirected] (95.center) to (96.center);
		\draw [style=black] (93.center) to (94.center);
		\draw [style=bluedirected] (100.center) to (99.center);
		\draw [style=greensolid] (101.center) to (102.center);
		\draw [style=greensolid] (102.center) to (83.center);
		\draw [style=greensolid] (83.center) to (84.center);
		\draw [style=greensolid] (84.center) to (101.center);
		\draw [style=greendirected] (104.center) to (105.center);
		\draw [style=black] (108.center) to (109.center);
		\draw [style=greensolid] (112.center) to (115.center);
		\draw [style=greendirected] (115.center) to (114.center);
		\draw [style=orangesolid] (121.center) to (120.center);
		\draw [style=orangesolid] (120.center) to (123.center);
		\draw [style=orangesolid] (121.center) to (122.center);
		\draw [style=orangesolid] (122.center) to (123.center);
		\draw [style=orangedirected] (124.center) to (125.center);
		\draw [style=black] (118.center) to (119.center);
		\draw (117.center) to (118.center);
		\draw (127.center) to (128.center);
		\draw [style=black] (128.center) to (129.center);
		\draw [style=bluesolid] (131.center) to (132.center);
		\draw [style=bluesolid] (133.center) to (132.center);
		\draw [style=bluesolid] (130.center) to (131.center);
		\draw [style=bluesolid] (130.center) to (133.center);
		\draw [style=orangedirected] (136.center) to (137.center);
		\draw [style=bluedirected] (134.center) to (135.center);
		\draw [style=greensolid] (145.center) to (148.center);
		\draw [style=greendirected] (148.center) to (147.center);
		\draw [style=greendirected] (148.center) to (150.center);
	\end{pgfonlayer}
\end{tikzpicture}
\caption{Teacher--Student, Deep
Probabilistic Logic (DPL), and Concordia architectures (left to right). Each couples
a neural model $\neural$ with a lifted graphical model $\logical$.}
\label{fig:composite-parallel}
\end{subfigure}

\caption{Regularisation-based composite systems; architectures
adapted from~\cite{DBLP:journals/corr/abs-2410-22077}.}
\label{fig:composite}
\end{figure}
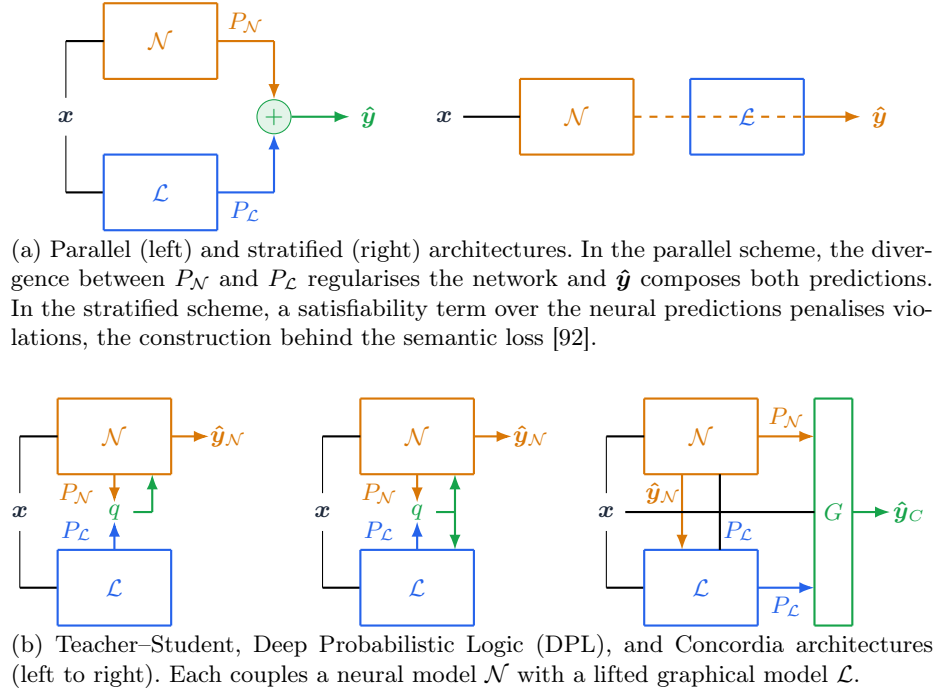

\begin{figure}[t]
\centering
\begin{tikzpicture}[
  mech/.style ={draw, line width=0.4pt, align=center, font=\scriptsize,
                inner xsep=4pt, inner ysep=5pt, minimum height=1.45cm},
  band/.style ={line width=0.4pt},
  tag/.style  ={font=\scriptsize, inner sep=1pt},
  link/.style ={line width=0.4pt, -{Stealth[length=1.6mm]}}]

\node[mech] (b1) {%
  \textbf{Exact WMC}\\[1pt]
  d-DNNF compilation\\
  DeepProbLog\\[1pt]
  \textit{$\#P$-hard}};
\node[mech, right=2.5mm of b1] (b2) {%
  \textbf{Deterministic approx.}\\[1pt]
  top-$k$ semirings\\
  Scallop\\[1pt]
  \textit{polynomial, biased}};
\node[mech, right=2.5mm of b2] (b3) {%
  \textbf{Stochastic approx.}\\[1pt]
  sampling or search\\
  WeightME, A-NeSI, ISED, ABL\\[1pt]
  \textit{unbiased, amortised, search-based}};

\draw[link] ([yshift=3mm]b1.north west) -- ([yshift=3mm]b3.north east);
\node[tag, anchor=south west] at ([yshift=3.6mm]b1.north west) {exact};
\node[tag, anchor=south east] at ([yshift=3.6mm]b3.north east) {approximate};
\path ([yshift=3.6mm]b1.north west) -- ([yshift=3.6mm]b3.north east)
  node[midway, anchor=south, font=\scriptsize\bfseries] {Inference exactness};

\draw[band] ([yshift=-5.5mm]b1.south west) rectangle ([yshift=-18mm]b3.south east);
\path ([yshift=-5.5mm]b1.south west) -- ([yshift=-5.5mm]b3.south east)
  node[midway, anchor=north, yshift=-1mm, font=\scriptsize\bfseries]
    {Execution target};
\path ([yshift=-18mm]b1.south west) -- ([yshift=-18mm]b3.south east)
  node[midway, anchor=south, yshift=1mm, font=\scriptsize, align=center]
    {Lobster: Datalog compiled to GPU kernels\\
     DOLPHIN: symbolic layer on CPU, probabilistic and gradient computation on GPU};

\draw[link] (b1.south) -- ([yshift=-5.5mm]b1.south);
\draw[link] (b2.south) -- ([yshift=-5.5mm]b2.south);
\draw[link] (b3.south) -- ([yshift=-5.5mm]b3.south);
\end{tikzpicture}
\caption{Inference exactness versus execution target.}
\label{fig:axis4}
\end{figure}
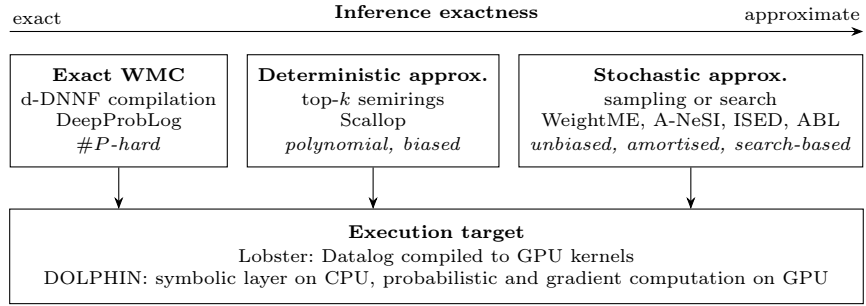

\subsection{Axis 4: Evaluation Mechanism}\label{sec:axis4}
The final axis analyses how programs are evaluated, based on how
the answer probability is obtained and on the target
hardware layer (Fig.~\ref{fig:axis4}).

What is computed ranges over a spectrum of exactness. At one end,
\emph{exact weighted model counting} sums the contribution of every
model. In practice, the program is compiled into a deterministic
decomposable negation normal form (d-DNNF) circuit on which weighted
counting is linear-time, the standard approach in the \system{ProbLog} family~\cite{DBLP:journals/tplp/FierensBRSGTJR15,DBLP:journals/ai/ManhaeveDKDR21}.
Exactness is costly, as WMC is $\#P$-hard and the circuit can be
exponential in the treewidth of the ground program. Approximation relaxes this in
two ways. \emph{Deterministic} approximation truncates the computation.
\system{Scallop}'s top-$k$ provenance semiring retains only the $k$ most likely
derivations of each tuple, reducing the exponential proof set to a
polynomial one at the cost of  a biased
estimate~\cite{DBLP:conf/nips/HuangLCSNSS21,DBLP:journals/pacmpl/LiHN23}.
\emph{Stochastic} approximation instead estimates the count or its
gradient by sampling. \system{WeightME}~\cite{DBLP:conf/icml/MaeneDR24} estimates the WMC gradient without bias and, under mild assumptions, attains probabilistic guarantees with logarithmically many SAT-oracle calls.
\system{A-NeSI}~\cite{vankrieken2023anesi} amortises inference, training a network to approximate the WMC.

ISED~\cite{ised} extends sampling to neural programs whose symbolic
component is a black box, estimating gradients from symbol assignments
sampled from the networks' outputs. Softened symbol grounding samples
symbol states from a Boltzmann distribution via projection-based
Markov chain Monte Carlo~\cite{li2023softened}, while abductive learning (ABL) searches for pseudo-labels that minimally
revise the networks' predictions to satisfy the knowledge
base~\cite{dai2019abductive}. ABL thus approximates the learning signal
through search rather than estimating the count itself. All three
operate only on the networks' outputs. 
\system{CLIPPER}~\cite{naik2026clipper}, a plug-in
compatible with each, exploits the latent representation space: an
integer linear program prunes abduced label combinations inconsistent
with representation-level similarity, guaranteeing that each sample
retains a logically consistent candidate.

Where it is computed is an orthogonal choice: the same degree of exactness can
target different layers, and GPU co-design scales evaluation without changing
what is computed. \system{Lobster}~\cite{DBLP:conf/asplos/Biberstein0DN26} compiles Datalog
evaluation to data-parallel GPU kernels through \system{Lobster}'s APM intermediate language,
whereas \system{DOLPHIN}~\cite{naik2025dolphin} keeps the symbolic layer on the
CPU and vectorises the probabilistic and gradient computation on the GPU. Since acceleration targets the hardware layer, it is orthogonal to the
exactness choice, although both systems currently pair it with
approximate inference. \system{KLay}~\cite{DBLP:conf/iclr/MaeneDM25} covers the exact side, restructuring
compiled arithmetic circuits into layered representations that
parallelise on GPUs.

The two also differ in the algebra they accelerate. \system{DOLPHIN} uses its own two vectorised provenances: \system{DAMP}, a fuzzy add-mult semantics, and \system{DTKP-AM}, an add-mult approximation of top-$k$ proofs~\cite{naik2025dolphin}. Across its benchmarks, \system{DAMP} is the faster option, and the more accurate one when the proof space outgrows top-$k$ truncation. \system{Lobster} likewise compiles its library of provenance semirings through the APM
abstraction~\cite{DBLP:conf/asplos/Biberstein0DN26}.

The mechanism also interacts with learning imbalances. Even under
balanced data, the symbolic program itself can induce large differences
in class-specific risks, and plain likelihood (WMC) maximisation does
not correct them. A more effective strategy is to estimate the marginal
of the hidden labels and enforce it as a constraint at training and
testing time~\cite{tsamoura2025imbalances} (see §\ref{sec:cross}).

Overall, scalability is determined by the inference mechanism rather
than by the language, since a program scales only if the chosen
approximation can capture its proof space.

\section{Corpus and Domain Analysis}
\label{sec:domains}

To ground the taxonomy, we organise a curated corpus of over 50
systems, methods, and applications, mostly from the past six years,
into four domains by primary research community: databases and
programming languages (DB/PL), core machine learning and AI, computer
vision, and robotics. Table~\ref{tab:domains} summarises representative
entries. AI dominates the corpus, DB/PL contributes much of the
theoretical core, and vision and robotics are less represented.
For six recurring scenarios, we then identify the language features
on which systems rely.

\begin{table}[t]
\centering
\caption{Representative domain summary of the surveyed corpus.}
\label{tab:domains}
\footnotesize
\setlength{\tabcolsep}{3pt}
\renewcommand{\arraystretch}{1.1}
\newlength{\roblabellen}
\settowidth{\roblabellen}{\footnotesize Robotics}
\newcommand{\domainlabel}[1]{\rotatebox[origin=c]{90}{#1}}
\newcommand{\blockstrut}{\rule[-0.28\roblabellen]{0pt}{0.58\roblabellen}}
\begin{tabular}{@{\hspace{3pt}}c@{\hspace{8pt}} >{\raggedright\arraybackslash}p{4.2cm} >{\raggedright\arraybackslash}p{6.8cm}@{}}
\hline\noalign{\smallskip}
 & \textbf{Formalism family} & \textbf{System/Approach} \\
\noalign{\smallskip}\hline\noalign{\smallskip}
\multirow{3}{*}{\domainlabel{DB/PL}}
 & Stratified Datalog
 & Scallop \cite{DBLP:journals/ftpl/LiHLN24,DBLP:journals/pacmpl/LiHN23},
   Lobster \cite{DBLP:conf/asplos/Biberstein0DN26} \\
 & Datalog theory \cite{DBLP:conf/pods/GreenKT07}
 & Datalog$^{\circ}$ \cite{DBLP:journals/jacm/KhamisNPSW24},
   trigger graphs \cite{DBLP:journals/pvldb/TsamouraCMU21} \\
 & Relational query DSL
 & TorchQL \cite{NaikOOPSLA2024TorchQL} \\
\noalign{\smallskip}\hline\noalign{\smallskip}
\multirow{9}{*}{\domainlabel{AI}}
 & Positive / chain-rule Datalog
 & Scallop~v0 \cite{DBLP:conf/nips/HuangLCSNSS21},
   TensorLog \cite{cohen2020tensorlog},
   NeuralLP \cite{yang2017neurallp},
   DRUM \cite{sadeghian2019drum},
   numerical NeuralLP \cite{WangICLR2020NeuralLPN},
   EduCe \cite{XuKBS2023EduCe},
   RNNLogic \cite{QuICLR2021RNNLogic} \\
 & Stratified Datalog
 & DOLPHIN \cite{naik2025dolphin},
   Vieira \cite{li2024vieira},
   ESCA \cite{HuangNeurIPS2025ESCA},
   DSR-LM \cite{ZhangACL2023DSRLM} \\
 & Probabilistic logic / WMC
 & DeepProbLog \cite{DBLP:journals/ai/ManhaeveDKDR21},
   A-NeSI \cite{vankrieken2023anesi},
   VAEL \cite{MisinoNeurIPS2022VAEL} \\
 & Stable-model ASP
 & NeurASP \cite{DBLP:conf/ijcai/YangIL20},
   SLASH \cite{skryagin2022slash},
   dPASP \cite{geh2024dpasp} \\
 & Prolog (SDCG / soft unif.)
 & DeepStochLog \cite{DBLP:conf/aaai/WintersMMR22},
   DeepSoftLog \cite{DBLP:conf/nips/MaeneR23} \\
 & ILP (diff.\ / symbolic)
 & $\partial$ILP \cite{evans2018dilp},
   $\alpha$ILP \cite{shindo2023alphailp},
   DiffLog \cite{si2019difflog},
   NeurRL \cite{GaoICLR2025NeurRL},
   DFORL \cite{GaoAIJ2024DFORL},
   MaxSynth \cite{HocquetteAAAI2024MDL} \\
 & Fuzzy FOL
 & LTN \cite{badreddine2022ltn},
   LNN \cite{sen2022lnn},
   NeuPSL \cite{pryor2023neupsl} \\
 & Markov logic
 & pLogicNet \cite{qu2019plogicnet},
   ExpressGNN \cite{zhang2020expressgnn} \\
 & Solver / black-box / circuits
 & ABL \cite{dai2019abductive},
   ISED \cite{ised},
   SATNet \cite{wang2019satnet},
   LLM rule extraction \cite{FangAAAI2024LLMNeSy},
   DeepLog \cite{derkinderen2025deeplog} \\
\noalign{\smallskip}\hline\noalign{\smallskip}
\multirow{4}{*}{\domainlabel{Vision}}
 & Functional DSL
 & NS-CL \cite{mao2019nscl},
   LEFT \cite{hsu2023left},
   NGS \cite{DBLP:conf/icml/LiHHCWZ20} \\
 & Temporal logic / relational
 & LASER \cite{huang2025laser},
   NSVS-TL \cite{ChoiECCV2024NSVS} \\
 & Abductive ASP / Boolean
 & online abduction \cite{DBLP:journals/ai/SuchanBV21},
   softened symbol grounding \cite{li2023softened},
   CLIPPER \cite{naik2026clipper} \\
 & Knowledge graph rules
 & NeSyVQA \cite{KhanAccess2025NeSyVQA} \\
\noalign{\smallskip}\hline\noalign{\smallskip}
\multirow{2}{*}{\domainlabel{Robotics}}
 & \blockstrut Temporal LTL/STL \cite{DBLP:conf/ijcai/GiacomoV13,DBLP:conf/formats/MalerN04}
 & formal NN training \cite{SunTRO2024LTLPlanning},
   pSTL-Diffusion \cite{MengRAL2024STLDiffusion} \\
 & \blockstrut OWL $+$ Datalog/SWRL \cite{horrocks2004swrl}
 & Deep-Onto \cite{NakawalaIJCARS2019DeepOnto} \\
\noalign{\smallskip}\hline
\end{tabular}
\end{table}

\paragraph{Visual perception and reasoning.}
Extracting symbolic structure from pixels and composing it with rules
often requires differentiable probabilistic reasoning with aggregation
and negation. \system{Scallop}~\cite{DBLP:journals/pacmpl/LiHN23} is
evaluated on MNIST-R, handwritten-formula recognition, \system{CLEVR},
and PacMan-Maze. As noted in §\ref{sec:taxonomy}, expressing the
\system{CLEVR} interpreter requires stratified negation and
aggregation~\cite{DBLP:journals/pacmpl/LiHN23}.

\paragraph{Spatio-temporal scene-graph learning.}
Extracting relations across video frames requires a temporal-%
relational specification language with differentiable probabilistic
reasoning. \system{LASER}~\cite{huang2025laser} pairs an LLM that extracts STSL
specifications, a neural model that predicts scene graphs, and a
differentiable alignment checker built on \system{Scallop}. The alignment loss turns the extracted specifications into a weak
supervision signal for the perception model.

\paragraph{Knowledge graph completion.}
Open-vocabulary knowledge-graph completion benefits from soft
unification or embedding-based equivalence.
\system{TensorLog}~\cite{cohen2020tensorlog} evaluates restricted chain-rule
Datalog differentiably, rule learners built on this scheme induce such
rules end-to-end~\cite{sadeghian2019drum,yang2017neurallp}, and
\system{DeepSoftLog}~\cite{DBLP:conf/nips/MaeneR23} adds unification in
embedding space.
Treating pairwise similarities as independent facts, however, does not
guarantee that the induced equality is transitive.
The equivalence semantics of Maene and
Tsamoura~\cite{maene2025equivalence} repairs this defect and
outperforms soft-unification baselines on standard benchmarks.

\paragraph{Natural-language reasoning.}
Here, a foundation model may be invoked as a component rather than
trained as a neural predicate through the rule layer.
\system{Vieira}~\cite{li2024vieira} exposes foundation models through
foreign predicates (§\ref{sec:axis3}) and is evaluated on language,
vision, and structured-data tasks. Related solver-backed LLM-logic
systems include \system{Logic-LM}~\cite{pan2023logicim} and
\system{SatLM}~\cite{ye2023satlm}.

\paragraph{Planning, program analysis, and bioinformatics.}
These scenarios require recursive Datalog over probabilistic or
large-scale fact bases. In \system{Scallop}'s
PacMan-Maze~\cite{DBLP:journals/pacmpl/LiHN23}, a neural network
extracts the grid contents from pixels and the rule layer computes
enemy-free reachable paths by transitive closure. Trained end-to-end
through the rule layer, the resulting agent outperforms purely neural
baselines. \system{Lobster}~\cite{DBLP:conf/asplos/Biberstein0DN26} extends
Datalog's classical
workloads~\cite{DBLP:journals/tkde/CeriGT89,DBLP:conf/rweb/Krotzsch25}
to the neurosymbolic setting. It compiles evaluation to GPU kernels via
the APM intermediate language and is experimentally evaluated at scale
on benchmarks including program reasoning, bioinformatics, and
planning.

\section{Cross-Cutting Analysis}
\label{sec:cross}

This section cuts across the axes: §\ref{ssec:inference-vs-training} covers uncertainty from
training to deployment and §\ref{ssec:coverage} maps scenarios to features.

\subsection{Inference versus Training}\label{ssec:inference-vs-training}

At inference, propagating uncertainty is optional. 
The Boolean ``neuro-then-symbolic'' strategy asserts neural
maximum-a-posteriori (MAP) predictions as symbolic input and runs a
fixed solver, such as least-fixpoint Datalog, to obtain one answer. For
function-free Datalog, data complexity is polynomial, but uncertainty
is discarded and an incorrect MAP fact cannot be revised.
Training removes
this freedom. A hard $\arg\max$ has zero gradient almost everywhere, so
the discarded uncertainty must be reintroduced, by
differentiating the soft semantics
directly~\cite{DBLP:journals/pacmpl/LiHN23,naik2025dolphin}, compiling
the constraint into a loss term~\cite{DBLP:conf/icml/XuZFLB18}, or
keeping a probabilistic objective without differentiating the
circuit, via sampling~\cite{DBLP:conf/icml/MaeneDR24},
amortisation~\cite{vankrieken2023anesi}, or a marginal constraint over the hidden labels~\cite{tsamoura2025imbalances}. 
Deploying the logic in Boolean form leaves two options: confine the relaxation to
training, as differentiable ILP does by extracting Boolean rules from a
$[0,1]$ relaxation~\cite{evans2018dilp,si2019difflog}, or keep the
logic off the gradient path, as in abductive
learning~\cite{dai2019abductive}.
\system{SATNet}~\cite{wang2019satnet} fits neither, as its
semidefinite relaxation of maximum satisfiability is the inference
mechanism at both training and test time.

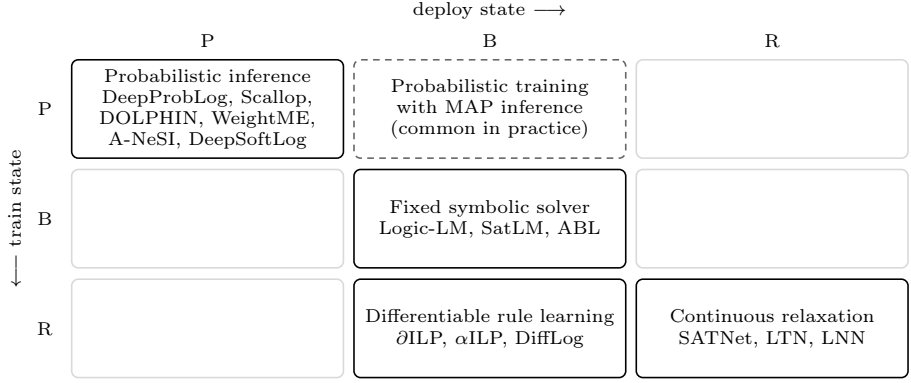
\begin{figure}[t]
\centering
\begin{tikzpicture}[
  cell/.style ={draw=black, line width=0.6pt, rounded corners=2.5pt, fill=white,
                align=center, font=\scriptsize,
                text width=3.45cm, minimum height=1.3cm, inner sep=2pt},
  used/.style ={cell, draw=black!60, dash pattern=on 2.5pt off 2pt},
  dead/.style ={cell, draw=black!15},
  pbr/.style  ={font=\scriptsize},
  axis/.style ={font=\scriptsize}]
\def\cA{0}\def\cB{3.73}\def\cC{7.46}
\def\rA{0}\def\rB{-1.45}\def\rC{-2.9}
\node[cell] at (\cA,\rA) {Probabilistic inference\\
  DeepProbLog, Scallop,\\ DOLPHIN, WeightME,\\ A-NeSI, DeepSoftLog};
\node[cell] at (\cB,\rB) {Fixed symbolic solver\\ Logic-LM, SatLM, ABL};
\node[cell] at (\cB,\rC) {Differentiable rule learning\\ $\partial$ILP, $\alpha$ILP, DiffLog};
\node[cell] at (\cC,\rC) {Continuous relaxation\\ SATNet, LTN, LNN};
\node[used] at (\cB,\rA) {Probabilistic training\\ with MAP inference\\ (common in practice)};
\node[dead] at (\cC,\rA) {}; \node[dead] at (\cA,\rB) {};
\node[dead] at (\cC,\rB) {}; \node[dead] at (\cA,\rC) {};
\node[pbr] at (\cA,0.9){P}; \node[pbr] at (\cB,0.9){B}; \node[pbr] at (\cC,0.9){R};
\node[axis] at (\cB,1.3) {deploy state $\longrightarrow$};
\node[pbr] at (-2.15,\rA){P}; \node[pbr] at (-2.15,\rB){B}; \node[pbr] at (-2.15,\rC){R};
\node[axis, rotate=90] at (-2.55,\rB) {$\longleftarrow$ train state};
\end{tikzpicture}
\caption{Train\,$\to$\,deploy grid (P: probabilistic, B: Boolean/MAP, R: relaxation).}
\label{fig:trajectory}
\end{figure}

Figure~\ref{fig:trajectory} classifies the surveyed families by their
uncertainty representation at training and deployment into \emph{probability
distribution} (P), \emph{Boolean/MAP value} (B), or \emph{continuous
relaxation}, i.e., $[0,1]$ values instead of Boolean ones (R).
P\,$\to$\,P systems (\system{DeepProbLog}~\cite{DBLP:journals/ai/ManhaeveDKDR21},
\system{Scallop}~\cite{DBLP:conf/nips/HuangLCSNSS21,DBLP:journals/pacmpl/LiHN23},
\system{DOLPHIN}~\cite{naik2025dolphin}, \system{WeightME}~\cite{DBLP:conf/icml/MaeneDR24},
\system{A-NeSI}~\cite{vankrieken2023anesi}\rev{, and \system{DeepSoftLog}~\cite{DBLP:conf/nips/MaeneR23}, whose soft
unification has possible-world
semantics}) retain an answer
distribution, so the rule layer can revise low-confidence neural
predictions, at the cost of intractable exact inference (§\ref{sec:axis4}).
B\,$\to$\,B systems (\system{Logic-LM}~\cite{pan2023logicim},
\system{SatLM}~\cite{ye2023satlm}, abductive learning
(ABL)~\cite{dai2019abductive}) keep fixed rules off every gradient
path, with purely Boolean inference and no probabilistic machinery.
NeuroLog-style abduction (Fig.~\ref{fig:abduction}), by contrast,
differentiates through the compiled proofs and belongs to the
P\,$\to$\,P family. R\,$\to$\,R systems
(\system{LTN}~\cite{badreddine2022ltn},
\system{LNN}~\cite{sen2022lnn}, \system{SATNet}~\cite{wang2019satnet}) 
stay differentiable throughout, but operate on relaxed truth values rather than on a distribution over possible worlds, so their output scores have no probabilistic interpretation.
Differentiable ILP
($\partial$\system{ILP}~\cite{evans2018dilp}, $\alpha$\system{ILP}~\cite{shindo2023alphailp}, \system{DiffLog}~\cite{si2019difflog};
R\,$\to$\,B) relaxes the rule space during training and deploys a purely symbolic program. The P\,$\to$\,B cell does not correspond to a dedicated system family
but to a common deployment practice, namely training through a
probabilistic objective and asserting only the MAP predictions at
inference, which trades the revision ability of the P column for
cheaper Boolean inference.

\subsection{Feature Coverage across Scenarios}\label{ssec:coverage}

Table~\ref{tab:decision} maps each scenario to its required features, and
no single dialect covers them all. Perception-driven tasks need
probabilistic semantics to propagate neural uncertainty into the rule
layer. Visual reasoning also needs aggregation and negation,
spatio-temporal tasks need temporal operators, knowledge graphs rely on
soft unification, and NL reasoning relies on foreign-function interfaces.

\begin{table}[t]
\centering
\caption{Mapping applications to language
features (\cmark\ required, \bmark\ beneficial).}
\label{tab:decision}
\footnotesize
\setlength{\tabcolsep}{4.5pt}
\renewcommand{\arraystretch}{1.1}
\begin{tabular}{@{}lcccccccc@{}}
\toprule
\textbf{Scenario} & \textbf{Prob.} & \textbf{Rec.} & \textbf{Neg.} &
\textbf{Agg.} & \textbf{Temp.} & \textbf{Soft} & \textbf{FFI} &
\textbf{Diff.}\\
\midrule
Visual reasoning  & \cmark & \bmark & \cmark & \cmark & --     & --     & --     & \cmark\\
Spatio-temporal   & \cmark & \cmark & \bmark & --     & \cmark & \bmark & --     & \cmark\\
Knowledge graphs  & \bmark & \cmark & --     & --     & --     & \cmark & --     & \cmark\\
NL reasoning      & \bmark & \bmark & \bmark & --     & --     & --     & \cmark & \bmark\\
Planning / RL     & \cmark & \cmark & --     & \bmark & --     & --     & --     & \cmark\\
Program analysis  & \bmark & \cmark & \bmark & \cmark & --     & --     & \bmark & \bmark\\
\bottomrule
\multicolumn{9}{@{}l@{}}{\scriptsize Prob.\ = probabilistic; Rec.\ =
recursion; Neg.\ = negation; Agg.\ = aggregation; Temp.\ = temporal;}\\
\multicolumn{9}{@{}l@{}}{\scriptsize Soft = soft unification; FFI =
foreign-function interface; Diff.\ = differentiable.}\\
\end{tabular}
\end{table}

\paragraph{Provenance semirings as the unifying abstraction.}
The provenance semiring (§\ref{sec:background}) is the common algebra underlying the surveyed systems. Changing the semiring changes the semantics without changing the
program, so one engine can
support several of the dialects that the decision matrix distinguishes. Concretely, a single
stratified Datalog engine with a differentiable semiring (\system{Scallop},
\system{Lobster}, \system{DOLPHIN}) already covers the visual-reasoning, planning, and
program-analysis rows of Table~\ref{tab:decision}. The temporal and soft-unification columns, however, have no obvious semiring formulation, the foreign-function column is orthogonal to the semiring choice, and probabilistic inference remains the scalability bottleneck (§\ref{sec:roadmap}).

\paragraph{Reliability and reasoning shortcuts.}
A NeSy model can satisfy the symbolic constraints while assigning
unintended meanings to the learned concepts, a \emph{reasoning
shortcut} that degrades generalisation and calibration.
Ensembling over models with conflicting interpretations exposes which
concepts are ambiguous and improves
reliability~\cite{DBLP:conf/uai/MarconatoBKVPT24}. Together with imbalance-aware training (§\ref{sec:axis4}),
these mechanisms are largely orthogonal to language design.
\rev{A companion survey systematises reasoning shortcuts, their
causes, and mitigations, including incorrect symbol grounding in
foundation models~\cite{marconato2025rs}. Wang et al.~\cite{wang2023multiinstance} delimit when the latent
concepts are identifiable at all under abduction-based training, a
precondition for any downstream reliability guarantee.
Concept bottleneck LLMs reveal the concepts the model
uses~\cite{sun2025cbllm}, so one can check its symbol grounding
directly.}

\section{Open Problems}
\label{sec:roadmap}

We identify five open problems, as stated below.

\paragraph{Can LLM-generated rule sets be statically analysed and repaired?}

For rule-based systems, the bottleneck is often writing the rules, and three couplings of LLMs with logic have emerged. In the
authoring direction, the LLM acts as an autoformaliser, by parsing
premises into logic programs or solver
inputs~\cite{wu2022autoformalization,olausson2023linc,gao2023pal,pan2023logicim,ye2023satlm}
or by generating answer set
programs~\cite{DBLP:conf/kr/IshayY023}. In the calling direction, the
model is a relation invoked from within the
rules~\cite{li2024vieira,huang2025laser}, and a rigorous semantics for
this abstraction, accounting for hallucination, non-determinism, and
cost, is itself open. In the constraining direction, logical
constraints are enforced on the LLM, at decoding time or through
fine-tuning~\cite{calanzone2025loco,zhang2023gelato,DBLP:conf/nips/ZhangKYB024}.
 
\paragraph{Can a dialect unify Boolean, probabilistic, fuzzy, and embedding semantics?}
The semantic families of §\ref{sec:taxonomy} are currently split across
separate languages. Provenance semirings unify several of these semantics (§\ref{sec:background}, §\ref{ssec:coverage}), and \system{DeepLog}~\cite{derkinderen2025deeplog} compiles specifications to parameterised algebraic circuits.
Unification at the
\emph{source} level remains open. It requires reconciling closed-world
Datalog with open-world soft
unification~\cite{DBLP:conf/nips/MaeneR23,maene2025equivalence} under
one type system, and identifying which semantics can be captured by a commutative semiring. Embedding equivalence and temporal operators lack a semiring formulation.

\paragraph{Can Datalog and ASP be bridged in one differentiable system?}
Stratified Datalog is a fragment of ASP, and the well-founded
semantics~\cite{DBLP:journals/jacm/GelderRS91} is a polynomial-time,
three-valued approximation of the stable model semantics that coincides
with it on stratified programs~\cite{DBLP:conf/iclp/GelfondL88},
whereas stable-model reasoning is intractable in general. A
differentiable engine spanning the hierarchy of
Fig.~\ref{fig:venn}a could therefore  let a single program move along the well-founded/stable axis, paying the higher inference cost only
where non-monotonic reasoning is required. The open question is whether
one relaxation can remain faithful to both the least-fixpoint and the
stable-model reading without committing to either at compile time.
 
\paragraph{How far can probabilistic inference scale?}
Exact WMC relies on compiled circuits whose size can be exponential in
the treewidth of the ground program (§\ref{sec:axis4}). Current systems either improve the
estimators, via unbiased gradient estimates with guarantees from logarithmically
many SAT calls~\cite{DBLP:conf/icml/MaeneDR24} or
amortised by a trained inference network~\cite{vankrieken2023anesi}, or compile evaluation to GPU
kernels~\cite{DBLP:conf/asplos/Biberstein0DN26,DBLP:conf/iclr/MaeneDM25,naik2025dolphin}. All return point
estimates without error bounds. Deriving computable bounds for top-$k$ truncation and sampling, e.g., from semiring convergence theory~\cite{DBLP:journals/jacm/KhamisNPSW24}, would make approximate inference certifiable for safety-critical deployment.
 
\paragraph{Can we verify neurosymbolic programs?}
A function-free rule layer is decidable and can, in principle, be
certified, with neural outputs treated as bounded or probabilistic
inputs. Trigger-graph
materialisation~\cite{DBLP:journals/pvldb/TsamouraCMU21}
provides completeness- and termination-preserving execution plans for
rule materialisation. Lifting it to the neurosymbolic setting means propagating input
intervals or label distributions through materialisation, bounding the
\emph{output} distribution rather than verifying a single run. A program is thus trustworthy only when its rules and concept groundings are certified together (§\ref{sec:cross}).

\section{Conclusion}
We examined rule-based languages for neurosymbolic AI along four axes:
semantics, expressiveness, neural integration, and evaluation
mechanism. We then reviewed state-of-the-art systems across four
domains, spanning databases and programming languages, machine
learning, vision, and robotics, and presented a decision matrix
mapping six application scenarios to the features they require. Our
analysis shows that no single dialect covers every scenario, with
provenance semirings coming closest to a common foundation. We close by identifying open
problems and future research directions that build on this declarative core.

\subsubsection*{\discintname}
 The authors have no competing interests to declare that are relevant to the content of this article.
 
\bibliographystyle{splncs04}
\bibliography{references}

\end{document}